\documentclass{article}
\usepackage{spconf,amsmath,graphicx}
\usepackage{xcolor}
\usepackage{microtype}
\usepackage{graphicx}
\usepackage{subfigure}
\usepackage{booktabs} 
\usepackage{multirow}
\usepackage[T1]{fontenc}
\usepackage{hyperref}
\usepackage{amsfonts}
\usepackage{fixltx2e}
\usepackage[utf8]{inputenc}
\usepackage{hhline}

\newcommand{\comment}[1]{}
\usepackage{algorithm}          
\usepackage{algpseudocode}      

\newcommand{\Loss}{\mathcal{L}}

\title{Knowledge Distillation for Multi-Target Domain Adaptation in Real-Time Person Re-Identification}
\name{\resizebox{\textwidth}{!}{Félix Remigereau, Djebril Mekhazni, Sajjad Abdoli, Le Thanh Nguyen-Meidine, Rafael M. O. Cruz and Eric Granger}}
\address{Laboratoire d'Imagerie, de Vision et d'Intelligence Artificielle (LIVIA) \\ Dept. of Systems Engineering, ETS Montreal, Canada}
\begin{document}
%
\maketitle
\begin{abstract}
Despite the recent success of deep learning architectures, person re-identification (ReID) remains a challenging problem in real-word applications. Several unsupervised single-target domain adaptation (STDA) methods have recently been proposed to limit the decline in ReID accuracy caused by the domain shift that typically occurs between source and target video data. Given the multimodal nature of person ReID data (due to variations across camera viewpoints and capture conditions), training a common CNN backbone to address domain shifts across multiple target domains, can provide an efficient solution for real-time ReID applications. Although multi-target domain adaptation (MTDA) has not been widely addressed in the ReID literature, a straightforward approach consists in blending different target datasets, and performing STDA on the mixture to train a common CNN. However, this approach may lead to poor generalization, especially when blending a growing number of distinct target domains to train a smaller CNN. 
To alleviate this problem, we introduce a new MTDA method based on knowledge distillation (KD-ReID) that is suitable for real-time person ReID applications. Our method adapts a common lightweight student backbone CNN over the target domains by alternatively distilling from multiple specialized teacher CNNs, each one adapted on data from a specific target domain. Extensive experiments\footnote{GitHub code: \href{https://github.com/fremigereau/MTDA\_KD\_REID}{https://github.com/fremigereau/MTDA\_KD\_REID.}}  conducted on several challenging person ReID datasets indicate that our approach outperforms state-of-art methods for MTDA, including blending methods, particularly when training a compact CNN backbone like OSNet. Results suggest that our flexible MTDA approach can be employed to design cost-effective ReID systems for real-time video surveillance applications.
\end{abstract}
%

\section{Introduction}\label{sec:intro}
\vspace*{-0.15cm}


Person ReID aims to recognize an individual captured over a set of non-overlapping camera viewpoints. It is an important function required in many computer vision applications, ranging from sports analytics to video surveillance \cite{he2020multi}. State-of-the-art approaches for person ReID are typically implemented with a deep learning (DL) model, e.g., deep Siamese networks, trained through metric learning to provide global appearance features. CNN backbones are often used to learn an embedding, where similar image pairs (with the same identity) are close to each other, and dissimilar image pairs (with different identities) are distant from each other. Popular supervised losses include cross-entropy, triplet, and contrastive losses based on positives and negatives pairs in a labeled dataset \cite{boudiaf2020unifying}. Despite their success, they remain complex models, and typically require optimizing many parameters using large annotated image datasets. They also suffer from poor generalization in the presence of domain shift, where the distribution of original video data captures from the source domain diverges w.r.t data from the operational target domain. Domain shifts are introduced by variations in capture conditions (i.e., pose and illumination), camera settings and viewpoints from which a person is observed, and lead to considerable changes in the distribution of images in different datasets \cite{mekhazni2020unsupervised, xuan2021intra}.


Given the cost of data annotation, several unsupervised STDA methods have recently been proposed to limit the decline of ReID accuracy caused by domain shifts. These methods seek to adapt CNNs trained with annotated source video data to perform well in a target domain by leveraging unlabeled data captured from that domain. To learn a discriminant domain-invariant feature representation from source and target data, STDA methods typically rely on discrepancy-based or adversarial approaches \cite{mekhazni2020unsupervised, wang2021knowledge, NGUYENMEIDINE2021104096, nguyen2020joint_DA, liu2020open, ekladious2020dual, nguyen2021incremental, NEURIPS2018_c5ab0bc6, khurana2021unsupervised, allen2020towards}. 

Although  MTDA based on unlabeled data from multiple targets is important in many real-world applications, it remains largely unexplored \cite{Nguyen-Meidine_2021_WACV}. Existing STDA techniques can be extended to MTDA by either adapting multiple models, one for each target domain, or by blending data from the multiple target domains, and then applying STDA on the mixture of target data. For image classification, an adversarial MTDA approach based on blending transfer recently proposed where all targets are viewed as one domain. In  \cite{chen2019blendingtarget}, target-specific representations were concatenated to deploy a common model \cite{gholami2018unsupervised}. Nevertheless, these approaches are either too complex (require one model per target domain), or generalize poorly on distinct target domains, particularly when adapting a smaller common CNN backbone on a growing number of targets. In \cite{NGUYENMEIDINE2021104096}, MTDA is performed by distilling information from target-specific teachers to a student model (deployed for testing), significantly reducing system complexity.

In person ReID, distributed video cameras may correspond to different target domains (defined by viewpoints and capture conditions), and the CNN backbone should therefore be adapted to generalize well across multiple target domains. Few MTDA methods are proposed for person ReID. Recently, Tian et al. \cite{tian2021camera} introduced an MTDA method for ReID. Inspired by intra-domain camera style, they propose a Camera Identity-guided Distribution Consistency method based on distribution consistency and discriminative embedding losses. The first loss aligns image style of source and targets through a generative approach, while the second predicts the camera for each identity to decrease distances for each individual across cameras. However, for real-time person ReID, specialized MTDA techniques are required to train smaller cost-effective CNN backbones that can address domain shifts across multiple data distributions. In this paper, we advocate for MTDA methods based on knowledge distillation (KD) \cite{wu2019distilled, Nguyen-Meidine_2021_WACV} to provide a better trade-off between CNN accuracy and efficiency. 

In this paper, we propose KD-ReID, a cost-effective MTDA method for real-time person ReID applications. A CNN backbone (teacher) is adapted for each target domain, using an STDA method, and the corresponding unlabeled target dataset. Then, the knowledge learned from all the target-specific teachers is alternatively distilled to a common lightweight CNN backbone (student). This paper expands significantly from the KD method in  \cite{Nguyen-Meidine_2021_WACV}, which was proposed for image classification tasks on larger CNNs. Additionally, KD-ReID differs considerably from a KD method for person ReID that was proposed in  \cite{wu2019distilled} for multi-source domain adaptation (MSDA). It was applied in a semi-supervised learning scenario and distills knowledge from multiple source models (teachers) to a target model (student) by aligning student-teacher similarity matrices. The accuracy and complexity of KD-ReID are compared extensively against state-of-the-art MTDA methods on several challenging person ReID datasets.

\section{Proposed Approach} \label{sec:Methodology}
\vspace*{-0.1cm}

\begin{figure}[h]
  \centering
  \subfigure[]{\includegraphics[scale=0.25]{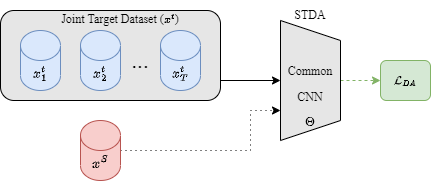}}\quad
  \subfigure[]{\includegraphics[scale=0.3]{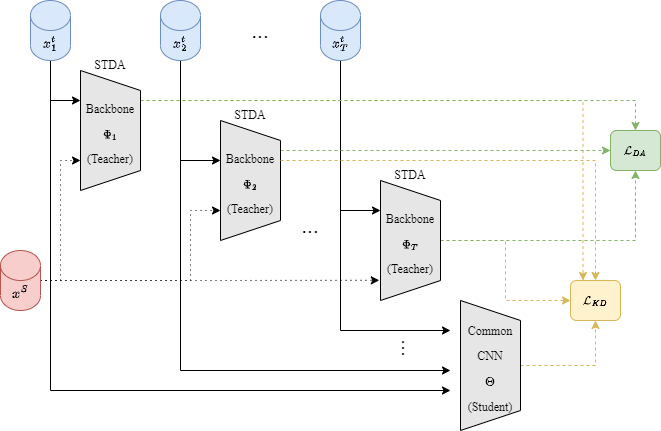}}
  \caption{Overview of MTDA methods. (a) Blending: target domain datasets are combined to form a dataset, and the  common CNN is adapted using a STDA method. (b) KD-ReID (ours): individual teachers are adapted using a STDA method for each target, and knowledge is distilled into the common CNN.}
  \label{fig:DA_gen_method}
  \vspace*{-0.1cm}
\end{figure}
\vspace*{-0.2cm}

Let $\mathbf{x}^s \in \mathcal{X}^s$ be the set of samples from the source domain, and ${y}^s \in \mathcal{Y}$ be their corresponding labels, where $\mathcal{X}^s$ is the source domain feature space, and $\mathcal{Y}$ is the source label space. Given the target domains feature space $\mathcal{X}^t$, let $\mathbf{x}^t = \left\{ \mathbf{x}_1^t, \mathbf{x}_2^t, ..., \mathbf{x}_T^t \right\} \in \mathcal{X}^t$ be the set of samples for $T$ unlabeled target domain datasets. We define the common CNN backbone as $\Theta$, and a set of target-specific CNN backbones, each one adapted to each target domain, as $\mathbf{\Phi} = \{\Phi_1, \Phi_2, ..., \Phi_T\}$. As illustrated in Figure \ref{fig:DA_gen_method} (a), a straightforward MTDA method for adapting a common CNN backbone $\Theta$ consists in blending the data $\mathbf{x}_i^t$ of all of the targets for $i = 1, 2, ..., T$, and then applying a STDA method on the resulting dataset $\mathbf{x}^t$.

We introduce KD-ReID, a MTDA method based on KD that is suitable for real-time person ReID. Our method adapts a common lightweight CNN backbone (student model) $\Theta$ to address multiple domain shifts, and perform well over all the target domains. As illustrated in Figure \ref{fig:DA_gen_method} (b), our proposed method distills knowledge learned by multiple specialized CNN backbones $\Phi_i$ (teacher models), each one adapted on data $\mathbf{x}_i^t$ from a specific target domain, for $i = 1, 2, ..., T$, using a STDA method. In particular, the KD-ReID method consists of three steps. \textbf{(1) Pre-training on source data:} All the CNN backbones (teachers $\Phi_i$ and student $\Theta$ models) undergo supervised pre-training on the labeled source domain dataset $\mathbf{x}^s$. This process should provide discriminant networks for person ReID on the source data. Any appropriate supervised loss functions for metric learning can be applied, including softmax cross-entropy \cite{loss_ces}, and hard samples mining triplet loss \cite{hermans2017defense}. \textbf{(2) STDA of teacher CNNs:} The teacher models $\Phi_i$ are adapted to their respective target domains using some STDA method with source $\mathbf{x}^s$ and target $\mathbf{x}_i^t$ datasets, for $i = 1, 2, ..., T$. Any STDA method can be applied, including the D-MMD \cite{mekhazni2020unsupervised}, and SPCL \cite{ge2020selfpaced} methods. \textbf{(3) KD to student CNN:} Knowledge from every teacher model $\Phi_i$ is distilled to the common student model $\Theta$ based on target datasets $\mathbf{x}_i^t$, for $i = 1, 2, ..., T$, and on the KD loss function, $\Loss_{KD}$ (described below). The KD-ReID method is summarized in Algorithm \ref{algorithm:D-MMD algorithm}.

\begin{algorithm}[!t]
    \begin{algorithmic}
        \caption{KD-ReID method.} 
        \label{algorithm:D-MMD algorithm}
        \Require labeled source datasets: $\mathbf{x}^s$, 
        unlabeled target data: $\mathbf{x}^t = \left\{ \mathbf{x}_1^t, \mathbf{x}_2^t, ..., \mathbf{x}_T^t \right\}$, 
        \State 1) Pre-train \textit{teacher} models, $\mathbf{\Phi} = \{\Phi_1, \Phi_2, ..., \Phi_T\}$, and \textit{student} model, $\Theta$, on the source labeled data $\mathbf{x}^s$
        \For{$i = 1, 2, ... T$}
                \For{each mini-batches $B^s \subset \mathbf{x}^s$ and $B^t \subset \mathbf{x}^t_i$}
                    \State 2) STDA: adapt \textit{teacher} $\Phi_i$ on source and target data
                \EndFor
        \EndFor
        \For{$i = 1, 2, ... T$}
                \For{each mini-batch $B^t \subset \mathbf{x}_i^t$}
                    \State 3) KD: adapt \textit{student} $\Theta$ using target data and $\Loss_{KD}$
                \EndFor
        \EndFor
    \end{algorithmic}
\end{algorithm}
\vspace*{-0.2cm}

Once the set of teacher models $\mathbf{\Phi}$ are adapted to respective target domains, KD is proposed to progressively integrate their knowledge into the common student model $\Theta$. Based on the self-similarity matrices of feature vectors from $\mathbf{\Phi}$  and $\Theta$, our proposed KD loss $\Loss_{KD}$ aligns student-teacher representations. Let $f_n = \Phi_i(x_n)$ be the feature vector output from the fully connected layer of $\Phi_i$ in response to sample $x_n$. Consider a feature matrix $\mathbf{F} = [f_1,f_2,...,f_N] \in \mathbb{R}^{D\times N}$ where $N$ is the batch size and $D$ is the dimension of feature vectors. 
The self-similarity matrix denoted as $\mathbf{A} \in \mathbb{R}^{N\times N}$ is defined as $a_{j,k} = \langle f_j,f_k \rangle$,  where $\langle \cdot, \cdot \rangle$ is the cosine distance between the feature vectors and $f_j$, $f_k \in \mathbf{F}$ .
Similarity matrices $\mathbf{A}^s$ and $\mathbf{A}^t$ are computed based on student $\mathbf{F}^s$ and teacher $\mathbf{F}^t$ feature matrices, respectively. Our KD loss $\Loss_{KD}$ is then computed as:
    $\Loss_{KD} = ||\mathbf{A}^s - \mathbf{A}^t||_F$, 
where $||\cdot,\cdot||_F$ is the Frobenius norm matrix. Optimizing $\Theta$ parameters with $\Loss_{KD}$ allows the student model to produce a distance matrix aligned to the teacher model. We only apply this loss on target samples $\mathbf{x}^t$. It allows tuning $\Theta$ parameters, while $\mathbf{\Phi}$ parameters are fixed. For each mini-batch $B^t$ from a given target domain $i$, we select only the corresponding teacher model $\Phi_i$. The order of targets for KD is selected randomly, and it changes at every epoch. 

An important benefit of the KD-ReID approach is its versatility. It is independent of the STDA methods used to adapt each $\Phi_i$. The choice of CNN backbones $\Phi_i$, and its STDA methods may differ, allowing to select the best configuration to address the specific challenges (i.e., domain shift) faced by each target domain. Moreover, KD-ReID allows distilling knowledge to a lighter common CNN backbone (student model), making it suitable for real-time applications.

\section{Experimental Results}\label{sec:Exp_methdology}
\vspace*{-0.15cm}

\noindent \textbf{Experimental Methodology:}
We evaluate MTDA methods across datasets, where each dataset is a target domain. Four benchmark datasets are used for person ReID experiments. CUHK03\cite{li2014cuhk03} consists of 13,164 images of 1,467 identities from 5 camera pairs. Market1501 \cite{zheng2015scalable} is comprised of a high number of images per identities (32,217 images of 1,501 identities) captured with 6 cameras. Compared to CUHK03, the bounding box annotations in Market1501 are lower quality, and persons captured in each box may be represented partially, which makes for a challenging dataset. DukeMTMC\cite{zhang2017multi} is designed such that each one of the 8 cameras contains unique identities. This dataset contains 36,441 images of 1,812 identities, which is similar in size to Market1501 in terms of the number of images and identities. Finally, MSMT17 \cite{wei2018msmt17} includes video from 15 indoor and outdoor cameras, and consists of 126,441 images of 4,101 identities. Since videos are captured in indoor/outdoor conditions, on different days, weather conditions, and times of the day, MSMT17 represents a more challenging and realistic dataset compared to the others. MSMT17 is always used as the source dataset to pre-train the CNNs. Given the diversity of capture conditions, there is a significant domain shift between images from these datasets \cite{mekhazni2020unsupervised}. 

ResNet50 \cite{he2016deep} is used to implement the target-specific CNN backbones (teachers), while OSNet\_x0\_25 \cite{zhou2019omni} implements the common CNN backbone. A fully connected layer with 2048 neurons is added to each backbone to output feature vectors. The original hyperparameter setup for D-MMD \cite{mekhazni2020unsupervised} and SPCL \cite{ge2020selfpaced} are used for our experiments. For KD-ReID, the Stochastic Gradient Descent (SGD) optimizer is used with a learning rate of 0.01. It is divided by 10 after every five epochs. A batch size of 64 is used, with 32 samples from the source and target data. Batches are constructed using groups of four images with the same identity corresponding to tracklets. Training is conducted until model accuracy improves by less than 0.5\% average mAP over five consecutive epochs.


\begin{table*}[!t]
\centering
\label{tab:overall_results}
\resizebox{0.86\textwidth}{!}{
\begin{tabular}{|l||c|c|c|c|c|c|c|c||c|c|} 
\hline
\multirow{3}{*}{\textbf{MTDA Method – Base STDA Method}} & \multicolumn{8}{c||}{\textbf{\textbf{Accuracy on Target Data (\%)}}}                                                                                                                      & \multicolumn{2}{c|}{\multirow{2}{*}{\textbf{Complexity}}}  \\ 
\cline{2-9}
                                                  & \multicolumn{2}{c|}{\textbf{Market1501}}     & \multicolumn{2}{c|}{\textbf{DukeMTMC}}       & \multicolumn{2}{c|}{\textbf{CUHK03}}         & \multicolumn{2}{c||}{\textbf{Average}}       & \multicolumn{2}{c|}{}                                      \\ 
\cline{2-11}
                                                  & \textbf{\textbf{mAP}} & \textbf{\textbf{R1}} & \textbf{\textbf{mAP}} & \textbf{\textbf{R1}} & \textbf{\textbf{mAP}} & \textbf{\textbf{R1}} & \textbf{\textbf{mAP}} & \textbf{\textbf{R1}} & \textbf{\# Parameters} & \textbf{FLOPs}                    \\ 
\hhline{|=::========::==|}
\textbf{Lower Bound: Superv. on Source Only}              & 24.9                  & 50.8                 & 27.9                  & 47.0                 & 25.6                  & 29.1                 & 26.1                  & 42.3                 & 0.4 M                  & 0.08 G                            \\ 
\hline
\textbf{One Model per Target – D-MMD (ResNet50)}           & 51.4                  & 74.9                 & 51.4                  & 69.3                 & 61.8                  & 65.9                 & 54.9                  & 70.0                 & $T$ x 27.7 M           & 2.70 G                            \\
\textbf{One Model per Target – D-MMD (OSNet\_x0\_25)}           & 35.9                  & 63.6                 & 37.1                  & 57.9                 & 40.9                  & 44.5                 & 38.0                  & 55.3                 & $T$ x 0.4 M           & 0.08 G                            \\
\textbf{Blending Targets – D-MMD}                 & 32.5                  & 59.5                 & 35.5                  & 55.6                 & 35.8                  & 40.5                 & 34.6                  & 51.9                 & 0.4 M                  & 0.08 G                            \\
\textbf{KD-ReID – D-MMD (Ours)}                   & 46.7                  & 70.9                 & 47.4                  & 66.8                 & \textbf{51.0}         & \textbf{55.3}        & 48.4                  & 64.3                 & 0.4 M                  & 0.08 G                            \\ 
\hline
\textbf{One Model per Target – SPCL (ResNet50)}            & 54.2                  & 75.3                 & 52.0                  & 69.6                 & 33.4                  & 34.8                 & 45.9                  & 59.9                 & $T$ x 27.7 M           & 2.70 G                            \\
\textbf{One Model per Target – SPCL (OSNet\_x0\_25)}           & 34.6                  &  58.0                & 31.9                  & 51.3                 & 21.1                  & 16.3                 &  29.2                 & 41.9                 & $T$ x 0.4 M           & 0.08 G                            \\
\textbf{Blending Targets – SPCL}                  & 31.5                  & 54.4                 & 31.5                  & 48.9                 & 16.4                  & 12.1                 & 26.5                  & 38.5                 & 0.4 M                  & 0.08 G                            \\
\textbf{KD-ReID – SPCL (Ours)}                    & 51.5                  & 74.6                 & 45.3                  & 64.7                 & 37.1                  & 39.8                 & 44.6                  & 59.7                 & 0.4 M                  & 0.08 G                            \\ 
\hline
\textbf{KD-ReID – Mixed D-MMD \& SPCL (Ours)}                   & \textbf{53.3}         & \textbf{75.4}        & \textbf{48.3}         & \textbf{67.0}        & 50.2                  & 54.4                 & \textbf{50.6}         & \textbf{65.6}        & 0.4 M                  & 0.08 G                            \\ 
\hline
\textbf{Upper Bound: Superv. Fine-Tuning on Targets}             & 62.3                  & 82.7                 & 58.3                  & 74.8                 & 62.6                  & 64.8                 & 61.1                  & 74.1                 & 0.4 M                  & 0.08 G                            \\
\hline
\end{tabular}
}
\caption{Performance of MTDA methods when \textbf{MSMT17} is used as the source dataset, and  \textbf{Market1501}, \textbf{DukeMTMC}, and \textbf{CUHK03} as target datasets ($T = 3$ targets), with 2 STDA techniques -- D-MMD and SPCL. For KD-ReID, ResNet50 implements the target-specific CNN backbones, and OSNet\_x0\_25 implements the common student CNN backbones. The lower bound performance is obtained through supervised training of OSNet\_x0\_25 on the labeled source dataset only, and the upper bound after supervised fine-tuning on blended target datasets. FLOPs are related to the extraction CNN features for one image sample.}
\label{tab:overall_results}
\end{table*} 

\noindent \textbf{Results and Discussion:}
Table \ref{tab:overall_results} compares the performance of KD-ReID against baseline methods in terms of the mean Average Precision (mAP), rank-1 accuracy, and the number of parameters and FLOPs. As expected, the lower-bound accuracy on target datasets is low due to the domain shift between the source and target data. Blending all the target data, and then applying an STDA method improves accuracy significantly over the lower bound. The proposed KD-ReID outperforms the blending methods on all datasets for both D-MMD and SPCL methods for STDA. Training a single model per target is a straightforward MTDA approach that achieves a higher level of accuracy, but incurs a high computational cost. As shown in Table \ref{tab:overall_results}, the SPCL technique for STDA produces a less accurate teacher model for the CUHK03 dataset than D-MMD. This observation motivates us to select the best STDA technique for each teacher model. The highest average accuracy is obtained when mixing STDA methods to adapt teachers for KD-ReID, highlighting its versatility. Table \ref{tab:overall_results} also shows the complexity of approaches in terms of the memory needed to store model weights (\# Parameters), and inference time for a sample (FLOPs). Note that our KD-ReID has significantly lower complexity than the teacher models from which it learns. Furthermore, although accuracy may decline as $T$ grows, the memory requirement for blending and KD-ReID is  independent of the number of target domains $T$.

Table \ref{tab:SOTA_compare} shows the performance of KD-ReID versus three related SOTA methods. First, Wu \textit{et. al.} \cite{wu2019distilled} proposed an MSDA method based on KD for a semi-supervised scenario. The method should be robust to domain shift across new domains since it benefits from some labeled target domain data, and the labeled data from multiple different source domains to adapt the common CNN backbone. In practice, this scenario is restrictive for ReID, given the need to collect datasets and annotate samples. Yet, our KD-ReID outperforms this method significantly 
, even with a lighter CNN backbone.
Second, Tian \textit{et. al.} \cite{tian2021camera} recently proposed the only MTDA method in literature for ReID. Our method significantly outperforms this approach with an improvement of 18.3\% in mAP, and 6.2\% in rank-1 on average, even using OSNet\_x0\_25, a much smaller CNN architecture (by a ratio close to 60). In contrast, our KD-ReID method divides the problem into simpler STDA problems, that may be solved using any combination of  STDA  methods, rather than resolving for the complex MTDA problem. Compression achieved with KD provides a lightweight student model to reproduce the performance of the highly optimized teachers. Recently, Li \textit{et. al.} \cite{li2019unsupervised} proposed a SOTA unsupervised person ReID method that relies on clustering. While this setting is more challenging, our method offers significantly better accuracy and is computationally efficient.
\begin{table}[!b]
    \resizebox{\columnwidth}{!}{
        \begin{tabular}{|l||cccccc|cc|}
            \hline
            \multirow{3}{*}{\textbf{\begin{tabular}[c]{@{}c@{}}MTDA Method\\ -- CNN Backbone\end{tabular}}} & \multicolumn{6}{c|}{\textbf{Accuracy on Target Data (\%)}} & \multicolumn{2}{c|}{\multirow{2}{*}{\textbf{Complexity}}} \\ \cline{2-7}
             & \multicolumn{2}{c|}{\textbf{Market1501}} & \multicolumn{2}{c|}{\textbf{DukeMTMC}} & \multicolumn{2}{c|}{\textbf{Average}} & \multicolumn{2}{c|}{} \\ \cline{2-9}
             & \multicolumn{1}{c|}{mAP} & \multicolumn{1}{c|}{R1} & \multicolumn{1}{c|}{mAP} & \multicolumn{1}{c|}{R1} & \multicolumn{1}{c|}{mAP} & R1 & \multicolumn{1}{c|}{\# Para.} & FLOPs \\ \hline \hline
            \textbf{Wu \cite{wu2019distilled} -- MobileNetV2} & \multicolumn{1}{c|}{33.5} & \multicolumn{1}{c|}{61.5} & \multicolumn{1}{c|}{29.4} & \multicolumn{1}{c|}{48.4} & \multicolumn{1}{c|}{31.5} & 55.0 & \multicolumn{1}{c|}{$T$ x 4.3 M} & 0.40 G \\ \hline
            \textbf{Tian \cite{tian2021camera} -- ResNet50} & \multicolumn{1}{c|}{35.9} & \multicolumn{1}{c|}{70.4} & \multicolumn{1}{c|}{33.6} & \multicolumn{1}{c|}{57.2} & \multicolumn{1}{c|}{34.8} & 63.8 & \multicolumn{1}{c|}{23.5 M} & 2.70 G \\ \hline
            \textbf{Li \cite{li2019unsupervised} -- ResNet50} & \multicolumn{1}{c|}{46.2} & \multicolumn{1}{c|}{69.2} & \multicolumn{1}{c|}{44.6} & \multicolumn{1}{c|}{62.3} & \multicolumn{1}{c|}{45,4} & 65.8 & \multicolumn{1}{c|}{$T$ x 23.5 M} & 2.70 G \\ \hline
            \textbf{\begin{tabular}[c]{@{}c@{}}KD-ReID with SPCL Teachers\\ -- OSNet\_x0\_25 (ours)\end{tabular}} & \multicolumn{1}{c|}{\textbf{54.2}} & \multicolumn{1}{c|}{\textbf{76.6}} & \multicolumn{1}{c|}{\textbf{47.8}} & \multicolumn{1}{c|}{\textbf{66.2}} & \multicolumn{1}{c|}{\textbf{51.0}} & \textbf{71.4} & \multicolumn{1}{c|}{\textbf{0.4 M}} & \textbf{0.08 G} \\ \hline
        \end{tabular}%
    }
    \caption{Performance of KD-ReID versus related SOTA methods on 2 target domains \textbf{MSMT17} $\rightarrow$ \textbf{Market1501} \& \textbf{DukeMTMC}. We consider multiple models for \cite{wu2019distilled} since it is a MSDA method, where $T$ is the number of targets.}     
    \label{tab:SOTA_compare}
\end{table}

\begin{figure}[!b]
\centering
\includegraphics[scale=0.25]{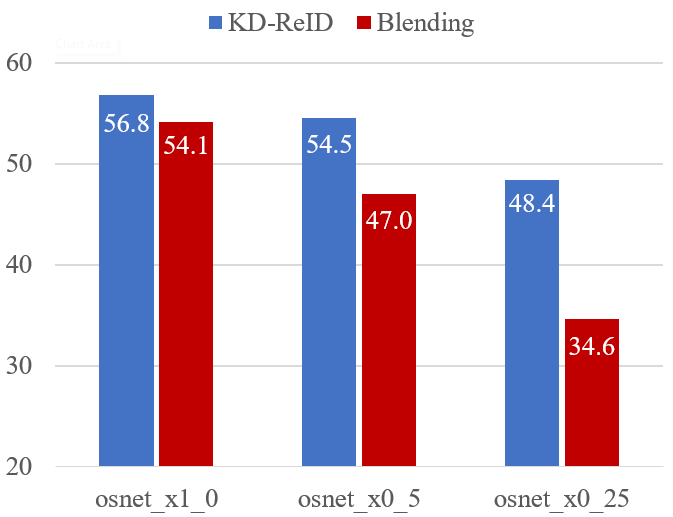}
\caption{Impact of student model complexity on the average mAP over targets (Market1501, DukeMTMC, and CUHK03), and using MSMT17 as source. The common CNN is OSNet, while teachers are ResNet50s, and D-MMD is used for STDA.}
\label{fig:complex_vs_map}
\end{figure}

Figure \ref{fig:complex_vs_map} compares the average performance of the common CNNs obtained using KD-ReID and blending. The experiments are conducted using three variations of the OSNet CNN: OSNet\_x1\_0, OSNet\_x0\_5, and OSNet\_x0\_25, having about 4.0M, 1.0M, and 0.4M parameters, respectively. Figure \ref{fig:complex_vs_map} shows that the gap in accuracy between methods widens as the model complexity is reduced. KD-ReID always outperforms blending, and remains more effective at adapting lightweight CNNs, thus more suitable for real-time applications.

\section{CONCLUSION} \label{sec:conclusion}
\vspace*{-0.15cm}

In this paper, KD-ReID is introduced as a versatile MTDA method for real-time person ReID. To adapt each teachers model, KD-ReID allows selecting the STDA method and CNN individually, and then knowledge of all teacher models is distilled to a common lightweight CNN (student). 
Our experiments conducted on challenging person ReID datasets indicate that student models adapted through KD-ReID outperform blending methods, and generalize well on all targets at the same time, even when adapting a small common CNN backbone on a growing number of targets. Such scalability paves the way for cost-effective systems for real-time applications.
 

\vspace*{0.2cm}
\noindent \textbf{Acknowledgements:} This work was supported by Nuvoola AI Inc., and the Natural Sciences and Engineering Research Council of Canada.

\bibliographystyle{IEEEbib}
\bibliography{refs}

\end{document}